\documentclass[10pt,twocolumn,letterpaper]{article}

\usepackage{cvpr}              

\usepackage[table,xcdraw,dvipsnames]{xcolor}

\newcommand{\name}[0]{UWAV}

\definecolor{cvprblue}{rgb}{0.21,0.49,0.74}
\usepackage[pagebackref,breaklinks,colorlinks,citecolor=cvprblue]{hyperref}

\usepackage{booktabs}
\usepackage{multirow}
\usepackage{makecell}
\usepackage{tabularx}
\usepackage{algorithm}
\usepackage{algpseudocode}
\usepackage{graphicx} 
\usepackage{amsfonts} 
\usepackage{amsmath}
\usepackage{amssymb}
\usepackage{array}
\usepackage{bbm}
\usepackage{pifont}
\usepackage{bbm}
\newcommand{\cmark}{\checkmark}%
\usepackage{soul}
\usepackage[accsupp]{axessibility}  
\usepackage{etoolbox}

\newcommand{\expnumber}[2]{{#1}\mathrm{e}{#2}}

\def\confName{CVPR}
\def\confYear{2025}

\title{UWAV: Uncertainty-weighted Weakly-supervised Audio-Visual Video Parsing}

\author{
Yung-Hsuan Lai$^{1,\dagger}$, Janek Ebbers$^{3}$, Yu-Chiang Frank Wang$^{1, 2}$, François Germain$^{3}$, \\
Michael Jeffrey Jones$^{3}$, Moitreya Chatterjee$^{3,\ddagger}$, \\
\normalsize \textsuperscript{1} Graduate Institute of Communication Engineering, National Taiwan University \quad \textsuperscript{2} NVIDIA, Taiwan \\
\normalsize \textsuperscript{3} Mitsubishi Electric Research Labs (MERL) \\
{\tt\small $^{\dagger}$r10942097@ntu.edu.tw  $^{\ddagger}$metro.smiles@gmail.com}
}

\begin{document}
\maketitle
\begin{abstract}
Audio-Visual Video Parsing (AVVP) entails the challenging task of localizing both uni-modal events (\ie, those occurring exclusively in either the visual or acoustic modality of a video) and multi-modal events (\ie, those occurring in both modalities concurrently). Moreover, the prohibitive cost of annotating training data with the class labels of all these events, along with their start and end times, imposes constraints on the scalability of AVVP techniques unless they can be trained in a weakly-supervised setting, where only modality-agnostic, video-level labels are available in the training data.
To this end, recently proposed approaches seek to generate segment-level pseudo-labels to better guide model training. However, the absence of inter-segment dependencies when generating these pseudo-labels and the general bias towards predicting labels that are absent in a segment limit their performance.
This work proposes a novel approach towards overcoming these weaknesses called Uncertainty-weighted Weakly-supervised Audio-visual Video Parsing (\name{}). Additionally, our innovative approach factors in the uncertainty associated with these estimated pseudo-labels and incorporates a feature mixup based training regularization for improved training. 
Empirical results show that \name{} outperforms state-of-the-art methods for the AVVP task on multiple metrics, across two different datasets, attesting to its effectiveness and generalizability. \footnote{Code is available at: \href{https://github.com/Franklin905/UWAV}{Code Link}.}
\end{abstract}    
\vspace{-10pt}
\section{Introduction}
\label{sec:intro}

Events that occur in the real world, often leave their imprint on the acoustic and visual modalities. Humans rely heavily on the synergy between their senses of sight and hearing to interpret such events. Audio-visual learning, which seeks to equip machines with a similar synergy, has emerged as one of the most important research areas within the multi-modal machine learning community. It aims to leverage both these senses (modalities) jointly, to enhance machine perception and understanding of real-world events. Various audio-visual learning tasks have been studied towards this end, including audio-visual segmentation~\citep{zhou2022audio, liu2024audio}, sound source localization~\citep{senocak2023sound, mo2023audio}, audio-visual event localization~\citep{tian2018audio, rao2022dual}, and audio-visual sound separation~\citep{chen2023iquery, ye2024lavss,chatterjee2022learning}.
However, many of these tasks assume that audio and visual streams would always be temporally aligned. This assumption often fails in real-world scenarios, where the sonic and visual imprints of events may not perfectly overlap.
For instance, one might hear an emergency siren approaching from a distance before it appears in the field of view.

\begin{figure}[t]
  \centering
  \includegraphics[width=\columnwidth]{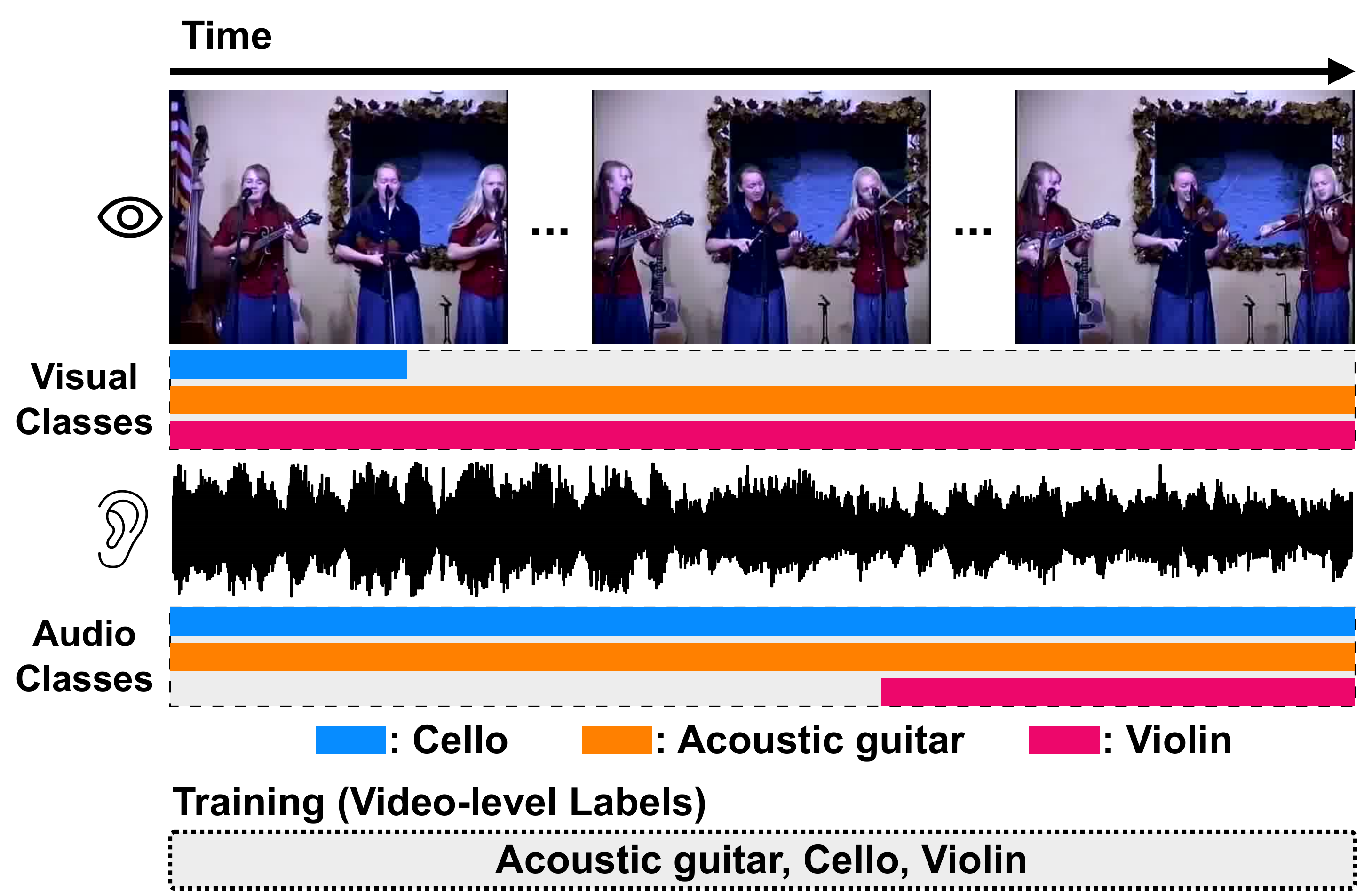}
  \caption{\textbf{A weakly-supervised AVVP task example.} Events, considered in this task, might be unimodal or multimodal. Even multimodal events, may not be temporally aligned in the audio and visual modalities, \eg the cello might only be visible in the first few seconds but might produce music, throughout the video.
  }
  \vspace{-10pt}
  \label{fig:teaser}
\end{figure}

In this work, in order to better understand the events occurring in a video, we explore the task of Audio-Visual Video Parsing (AVVP)~\cite{tian2020unified}. Its goal is to recognize and localize all audio, visual, and audio-visual events occurring in the video. See Figure~\ref{fig:teaser} for an example of this task. The task setup is to perform this prediction for every one-second segment of a video. 
This task poses two principal challenges, from a machine learning standpoint: (i) The audio and visual events, that occur, might not be temporally aligned, \eg if an event becomes audible before its source enters the camera field of view, or the sound source is not visible at all, and (ii) due to the high costs of annotating video segments with per-segment labels, only modality-agnostic, video-level labels are provided during training, \ie, these labels specify which events occur in a video but lack details about the segments or the modality in which they occur.

Prior works in the area can be grouped into two orthogonal research directions. The first focuses on enhancing model architectures~\citep{mo2022multi, yu2022mm, zhou2024label}.
Despite advancements in this direction, the absence of fine-grained labels to guide the model during training continues to pose an impediment towards the generalizability of such models. As a result, recent approaches have focused on the second direction of research which aims at generating richer pseudo-labels for improved training, either at the video-level~\citep{wu2021exploring, cheng2022joint} or segment-level~\citep{zhou2023improving, fan2023revisit, lai2023modality, rachavarapu2024weakly}. In particular, \citet{rachavarapu2024weakly} propose prototype-based pseudo-labeling (PPL), which seeks to train a pseudo-label generation module in conjunction with a core inference module. However, due to the lack of sufficient training data, this method struggles to generalize. 
On the other hand, VPLAN~\citep{zhou2023improving}, VALOR~\citep{lai2023modality}, and LSLD~\citep{fan2023revisit} leverage large-scale pre-trained foundation models, such as CLIP~\cite{radford2021learning} and CLAP~\cite{wu2023large}, along with ground-truth video-level labels to generate segment-level pseudo-labels for each of the two modalities. Audio/Visual segments (\eg the audio corresponding to the segment in question and the visual frame at the center of the segment) are fed into CLAP/CLIP, one segment at a time, to generate these pseudo-labels. Despite the significant improvement that these pseudo-label generation methods achieved, the correctness of the generated labels is still limited, constrained primarily by the lack of understanding of inter-segment dynamics. For instance, if a crowd is \emph{cheering} in a segment of the video, it is more likely that the crowd might also be \emph{clapping} right before or after.

To address the oversight of inter-segment dependencies and other shortcomings in existing pseudo-label generation methods, we introduce a novel, uncertainty-based, weakly-supervised, video parsing model called \name{} (\underline{U}ncertainty-weighted \underline{W}eakly-supervised \underline{A}udio-visual \underline{V}ideo Parsing), capable of generating improved segment-level pseudo-labels for better training of the inference module.
We resort to transformer modules~\cite{vaswani2017attention} to equip \name{} with the ability to capture temporal relationships between segments and pre-train it on a large-scale, supervised audio-visual event localization dataset~\cite{geng2023dense}. Subsequently, this pre-trained model is used to generate segment-level pseudo-labels for each modality on a target, small-scale dataset which only provides weak (\ie, video-level) supervision. Such a design permits a more holistic understanding of the video, resulting in more accurate pseudo-labels. 
Additionally, \name{} factors in the uncertainty associated with these estimated pseudo-labels in its optimization. That uncertainty is the result of the shift in the domain of the target dataset, insufficient model capacity, \etc and is computed at training time as the confidence scores associated with these labels.
To further enhance the model’s ability to learn in this small-scale, weakly-supervised data regime, we also employ a feature mixup strategy. This approach adds more regularization constraints by training on mixed segment features alongside interpolated pseudo-labels, which not only lessens the influence of noise but also enriches the training data, thereby reducing overfitting. Moreover, \name{} addresses a critical class imbalance issue in the pseudo-label enriched training data, viz. most event classes in any given segment of a video are absent/negative (\ie, they do not occur), while only a handful of them do. This creates a natural bias in the training set, making it difficult to learn the positive events. To counter this, we propose a class-frequency aware re-weighting strategy that lays greater emphasis on the accurate classification and localization of positive events. 
By incorporating these components into its design, our proposed method (\name{}) outperforms competing state-of-the-art approaches across two different datasets, viz. Look, Listen, and Parse (LLP)~\cite{tian2020unified} and the Audio-Visual Event Localization (AVE)~\cite{tian2018audio}, on multiple evaluation metrics.

In summary, our contributions are the following:
\begin{itemize}
    \item We introduce a novel, weakly-supervised method called \name{}, capable of synthesizing temporally coherent pseudo-labels for the AVVP task.
    \item To the best of our knowledge, ours is the first method for the AVVP task, which factors in the uncertainty associated with the estimated pseudo-labels while also regularizing it with a feature mixup strategy.
    \item \name{} outperforms competing state-of-the-art approaches for the AVVP task, across two different datasets on multiple metrics which attest to its generalizability.
\end{itemize}
\section{Related Works}
\vspace{-15pt}
\noindent \paragraph{Audio-Visual Learning:}
Audio-visual learning has emerged as an area of active research, aiming to develop models that synergistically integrate information from both audio and visual modalities for improved perception and understanding. 
Towards this end, various audio-visual tasks have been explored by the community so far, such as audio-visual segmentation~\citep{zhou2022audio, mao2023multimodal, yang2024cooperation, liu2024audio}, sound source localization~\citep{hu2022mix, huang2023egocentric, senocak2023sound, mo2023audio}, event localization~\citep{tian2018audio, zhou2021positive, rao2022dual}, navigation~\citep{chen2020soundspaces, chen2020learning, chen2021semantic, younes2023catch, yu2022sound, majumder2021move2hear}, generation~\citep{parida2022beyond, ruan2023mm, xing2024seeing}, question answering~\citep{geng2021dynamic, yun2021pano, li2022learning, shah2022audio}, and sound source separation~\citep{chatterjee2021visual, tian2021cyclic, chen2023iquery, ye2024lavss}. In this work, we focus on the task of audio-visual video parsing (AVVP) where the goal is to temporally localize events occurring in a video. Unlike many other audio-visual learning tasks, AVVP does not assume that events are always aligned across modalities. Some events could be exclusively uni-modal while others may have an audio-visual signature, which requires complex reasoning.

\noindent \paragraph{Audio-Visual Video Parsing (AVVP):}
To address the challenges of the AVVP task~\citep{tian2020unified,lai2023modality,rachavarapu2024weakly},~\citet{tian2020unified} proposed a Hybrid Attention Network (HAN) and a learnable Multi-modal Multiple Instance Learning (MMIL) pooling module. The HAN model facilitates the exchange of information within and across modalities using self-attention and cross-attention layers, while the MMIL pooling module aggregates segment-level event probabilities from both modalities to produce video-level probabilities.
Building on this foundation, recent works advanced the field from the following two perspectives.
The first group of studies~\citep{mo2022multi, yu2022mm, zhou2024label} focuses on enhancing model architectures. In particular, \citet{mo2022multi} proposed the Multi-modal Grouping Network (MGN) to explicitly group semantically similar features within each modality to improve the reasoning process, while \citet{yu2022mm} proposed the Multi-modal Pyramid Attentional Network (MM-Pyramid) to capture events of varying durations by extracting features at multiple temporal scales. Our proposed method is orthogonal to this line of research and can be integrated with any of these backbones.

The second direction focuses on generating pseudo-labels for improved training, either at the video-level~\citep{wu2021exploring, cheng2022joint} or the segment-level~\citep{zhou2023improving, fan2023revisit, lai2023modality, rachavarapu2024weakly}. VPLAN~\citep{zhou2023improving}, VALOR~\citep{lai2023modality}, and LSLD~\citep{fan2023revisit} utilize pre-trained CLIP~\cite{radford2021learning} and CLAP~\cite{wu2023large} along with ground-truth video-level labels to predict pseudo-labels for each modality on a per-segment basis. In contrast, PPL~\citep{rachavarapu2024weakly} uses the HAN model itself to generate pseudo-labels by constructing prototype features for each class and assigning labels to each segment based on the similarity between its feature and the prototype features.
While these pseudo-label generation methods have substantially improved model performance on the AVVP task, they still exhibit some limitations. For instance, to derive accurate pseudo-labels, PPL might require a large enough training set to learn good prototype features, which might pose challenges when applied to smaller datasets. Our proposed method overcomes this problem.
On the other hand, methods that leverage CLIP and CLAP to generate pseudo-labels often ignore temporal relationships between segments or the uncertainty associated with these labels. Our work also seeks to plug this void.
\begin{figure*}[t]
  \centering
  \includegraphics[width=1.0\textwidth]{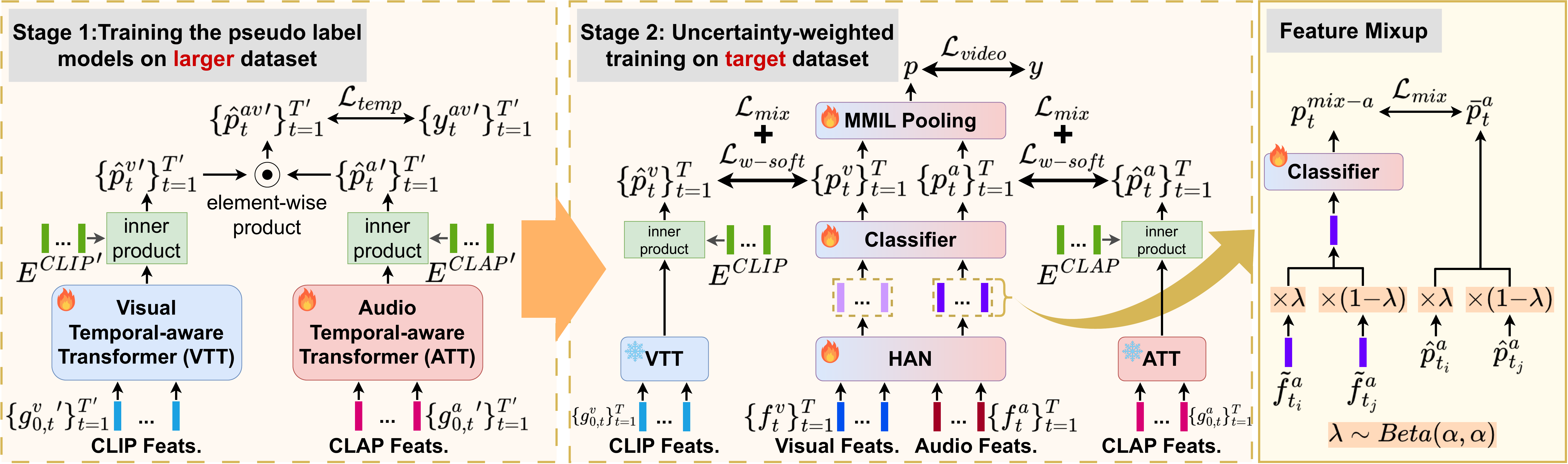}
  \caption{\textbf{\name{} framework:} In stage 1, pseudo-label generation modules are equipped with the ability to capture temporal relationships between segments by pre-training on a large-scale, supervised, audio-visual event localization dataset. In stage 2, temporally coherent, uncertainty-weighted pseudo-labels, derived from the pre-trained pseudo-label generation module, are used to guide the learning of the inference model (HAN) aided by a class-balanced loss re-weighting and uncertainty-weighted feature mixup strategy. Note that we use the feature mixup strategy in both modalities while we only show the breakdown of the mixup operation for the audio modality.
  }
  \label{fig:main_figure}
  \vspace{-10pt}
\end{figure*}

\section{Preliminaries}

\paragraph{Problem Formulation:}
The AVVP task~\citep{tian2020unified} aims to localize all visible and/or audible events in each one-second segment of a video. Specifically, an audible video is split into $T$ one-second segments, denoted as $\{V_t, A_t\}_{t=1}^T$. Each segment is annotated with a pair of ground-truth labels $y^v_t \in \{0, 1\}^C, y^a_t \in \{0, 1\}^C$, where $y^v_t$ denotes visual events, $y^a_t$ denotes audio events, and $C$ denotes the total number of events in the pre-defined event set of the data. However, owing to the weakly-supervised nature of the task setup $(y^v_t, y^a_t)$ are unavailable during training. Instead, only the modality-agnostic, video-level labels $y\in \{0, 1\}^C$ are available, where $1$ indicates the presence of an event at any time (either in the audio or visual stream or both) while $0$ indicates an event's absence in the video.

\paragraph{Pseudo-Label Based AVVP Framework:}
The Hybrid Attention Network (HAN)~\citep{tian2020unified} is a commonly used model for the AVVP task. The model works by first utilizing pre-trained visual and audio backbones to extract features from the visual and audio segments respectively, which are then projected to two $d$-dimensional feature spaces. The resulting visual segment-level features are denoted by $F^v = \{f^v_t\}^T_{t=1}\in \mathbb{R}^{T\times d}$, while the audio segment-level features are denoted by $F^a = \{f^a_t\}^T_{t=1}\in \mathbb{R}^{T\times d}$. These features are provided as input to the HAN model. In the model, information across segments within a modality and across modalities is exchanged through self-attention and cross-attention layers, as shown below:
\begin{equation}
\Tilde{f}^v_t = f^v_t + \underbrace{\text{Attn}(f^v_t, F^v, F^v)}_{\text{Self-Attention}} + \underbrace{\text{Attn}(f^v_t, F^a, F^a)}_{\text{Cross-Attention}},
\end{equation}

\vspace{-15pt}

\begin{equation}
\Tilde{f}^a_t = f^a_t + \underbrace{\text{Attn}(f^a_t, F^a, F^a)}_{\text{Self-Attention}} + \underbrace{\text{Attn}(f^a_t, F^v, F^v)}_{\text{Cross-Attention}},
\end{equation}
where $\text{Attn}(Q,K,V)$ denotes the standard multi-head attention mechanism~\citep{vaswani2017attention}.
Finally a classifier, shared across both modalities, transforms the visual segment-level features $\Tilde{F}^v = \{\Tilde{f}^v_t\}^T_{t=1}\in \mathbb{R}^{T\times d}$ (\textit{resp.} audio segment-level features $\Tilde{F}^a = \{\Tilde{f}^a_t\}^T_{t=1}$) into visual segment-level logits $\{z^v_t\}^T_{t=1} \in \mathbb{R}^{T\times C}$ (\textit{resp.} audio segment-level logits $\{z^a_t\}^T_{t=1}$). Segment-level probabilities $\{p^v_t\}_{t=1}^T, \{p^a_t\}_{t=1}^T \in \mathbb{R}^{T\times C}$ are then obtained by applying the sigmoid function on $\{z^v_t\}^T_{t=1}$ and $\{z^a_t\}^T_{t=1}$.

Since, only video-level labels $y$ are available during training, \citet{tian2020unified} introduce an attentive MMIL pooling module to learn to predict video-level probabilities $p\in \mathbb{R}^C$:
\vspace{-5pt}
\begin{equation}
W_{modal}^{v,a} = \text{Softmax}_{modal}(\text{FC}_{modal}(\Tilde{F}^{v,a})),
\end{equation}

\vspace{-10pt}

\begin{equation}
W_{time}^{v,a} = \text{Softmax}_{time}(\text{FC}_{time}(\Tilde{F}^{v,a})),
\end{equation}
where $\text{FC}_{modal}$ and $\text{FC}_{time}$ are two learnable fully-connected layers, $\Tilde{F}^{v,a}=\text{Stack}(\Tilde{F}^{v}, \Tilde{F}^{a})\in \mathbb{R}^{2\times T\times d}$ denotes the stacked visual and audio features along the first dimension, $\text{Softmax}_{modal}(\cdot)$ denotes the softmax operation along the modality dimension (\ie, across $v, a$), while $\text{Softmax}_{time}(\cdot)$ denotes the softmax operation along the temporal dimension (\ie, across $1,\dots, T$).
Video-level probabilities $p\in \mathbb{R}^C$ are then obtained via:
\vspace{-5pt}
\begin{equation}
p = \sum_{m=\{v,a\}} \sum_{t=1}^T W^{m,t}_{modal} \odot W^{m,t}_{time} \odot p^m_t,
\end{equation}
where $\odot$ denotes the element-wise product. The HAN model is then optimized with the binary cross-entropy (BCE) loss between the estimated video-level probabilities $p$ and video-level labels $y$:
$\mathcal{L}_{video} = \text{BCE}(p, y)$.
\section{Proposed Approach}
In this section, we detail our proposed approach (\name{}). At a high level, \name{} works by generating better segment-level pseudo-labels to improve the training of a multi-modal transformer-based inference module, \eg HAN. Moreover, \name{} factors in the uncertainty associated with these pseudo-labels, addresses the imbalance in the training data, and introduces self-supervised regularization constraints, which all lead to better performance. Figure~\ref{fig:main_figure} shows an overview of our proposed framework.

\subsection{Temporally-Coherent Pseudo-Label Synthesis}
One major issue that plagues prior works, based on pseudo-label generation~\citep{zhou2023improving, lai2023modality, fan2023revisit}, is that they do not capture the temporal dependencies between neighboring segments when generating the pseudo-labels. That is, the generated pseudo-labels are not temporally coherent.
To plug this void, we propose to incorporate transformer modules~\cite{vaswani2017attention} into the pseudo-label generation pipeline, which maps CLIP/CLAP encodings of a segment's visual frame/audio to pseudo-labels. Specifically, two separate transformers are introduced, one each for the visual/audio pseudo-label synthesis modules.

\noindent \textbf{Pre-Training:}
Training transformers often requires sufficiently large training data while datasets commonly used for the weakly-supervised AVVP task are relatively small.
To mitigate this challenge, we propose to first pre-train the transformer-equipped pseudo-label generation module on a large-scale, supervised, audio-visual event localization dataset -- the UnAV~\citep{geng2023dense} dataset.
Specifically, given an audible video of duration $T'$ seconds from the pre-training dataset, we split the video into $T'$ one-second segments $\{V'_t, A'_t\}_{t=1}^{T'}$, with corresponding audio-visual event labels ${y^{av}_t}'\in \{0, 1\}^{C'}$, where $1$ indicates the presence of an event in both modalities and $0$ its absence in at least one modality, while $C'$ denotes the total number of event classes in the pre-training dataset.
Next, the video frame at the temporal center of the visual segment is transformed into visual features ${G^v_0}' = \{ {g^v_{0,t}}{}' \}_{t=1}^{T'} \in \mathbb{R}^{T'\times d_1}$ with CLIP's~\cite{radford2021learning} image encoder.
These features are then fed into the corresponding transformer of the visual stream, consisting of $L$ encoder blocks, each block containing a self-attention layer, LayerNorm~\citep{ba2016layer} (LN), and a 2-layer feed-forward network (FFN):
\vspace{-14pt}
\begin{align}
{\Tilde{G}^{v}_{l}}{}' &= \text{LN}({G^v_l}' + \text{Attn}({G^v_l}', {G^v_l}', {G^v_l}')), \\
{G^v_{l+1}}' &= \text{LN}({\Tilde{G}^{v}_{l}}{}' + \text{FFN}({\Tilde{G}^{v}_{l}}{}') ).
\end{align}
Concurrently, we convert each event category label in the pre-training dataset into a textual event feature ${e^{\textit{CLIP}}_{c}}'\in \mathbb{R}^{d_1}$ by filling in the pre-defined caption template: ``A photo of $<$EVENT NAME$>$'' with the corresponding event name and passing it to CLIP's text encoder.
Equipped with the visual segment-level features ${G^{v}_{L}}{}' = \{ {g^v_{L,t}}{}' \}_{t=1}^{T'} \in \mathbb{R}^{T'\times d_1}$ and the textual event features ${E^{\textit{CLIP}}}'=\{ {e^{\textit{CLIP}}_{c}}' \}_{c=1}^{C'} \in \mathbb{R}^{C'\times d_1}$, we derive visual segment-level logits ${\hat{z}^{v}_t}{}' \in \mathbb{R}^{C'}$ and probabilities ${\hat{p}^{v}_t}{}'$ as follows:
\vspace{-5pt}
\begin{equation}
{\hat{p}^{v}_t}{}' = \text{Sigmoid}({\hat{z}^{v}_t}{}'),\ {\hat{z}^{v}_t}{}' = {{E^{\textit{CLIP}}}'} \cdot {g^v_{L,t}}{}'^\top.
\end{equation}

Similar operations are performed in the audio pseudo-label generation pipeline. The raw waveforms corresponding to the $1$-second audio segments are transformed into audio features ${G^a_0}' \in \mathbb{R}^{T'\times d_2}$ with CLAP's~\cite{geng2023dense} audio encoder and fed into the corresponding transformer consisting of $L$ encoder blocks.
Correspondingly, the textual event features ${E^{\textit{CLAP}}}' \in \mathbb{R}^{C'\times d_2}$ are generated with the caption template: ``This is the sound of $<$EVENT NAME$>$'' by passing it through CLAP's text encoder. Audio segment-level logits ${\hat{z}^{a}_t}{}' \in \mathbb{R}^{C'}$ and probabilities ${\hat{p}^{a}_t}{}'$ can then be derived in the same manner:
$
{\hat{p}^{a}_t}{}' = \text{Sigmoid}({\hat{z}^{a}_t}{}'), \ {\hat{z}^{a}_t}{}' = {{E^{\textit{CLAP}}}'} \cdot {g^a_t}{}'^\top.
$

Since the events occurring in the pre-training dataset (UnAV) are audio-visual, we multiply the segment-level visual and audio event probabilities to enforce the predicted labels to be multi-modal in nature: $\{ {\hat{p}^{av}_t}{}' \}_{t=1}^{T'}\in \mathbb{R}^{T'\times C'}$. This network is then trained with the binary cross-entropy (BCE) loss:
\begin{equation}
\mathcal{L}_{temp} = \text{BCE}({\hat{p}^{av}_t}{}', {y^{av}_t}'),\ {\hat{p}^{av}_t}{}' = {\hat{p}^{v}_t}{}'\odot {\hat{p}^{a}_t}{}'.
\end{equation}

\noindent \textbf{Pseudo-Label Generation on Target Dataset:}
With the pre-trained pseudo-label generation modules in place, we proceed to employ them for the pseudo-label generation process in the target dataset for the AVVP task. 
Specifically, the center frame of each of the visual segments $\{V_t\}_{t=1}^T$ of the target dataset are passed into CLIP's image encoder, whose output is passed into the pre-trained visual transformer to generate segment features $G^{v}_{L} = \{ g^v_{L,t} \}_{t=1}^T \in \mathbb{R}^{T\times d_1}$. At the same time, the caption template: ``A photo of $<$EVENT NAME$>$'' is used to obtain textual features corresponding to each of the event classes in the target dataset for the AVVP task: $E^{\textit{CLIP}} \in \mathbb{R}^{C\times d_1}$. Segment-level visual logits $\hat{z}^{v}_t\in \mathbb{R}^{C}$ can be derived by computing their inner product.
We also pre-define class-wise visual thresholds $\theta^v\in \mathbb{R}^C$ to transform segment-level visual logits into binary pseudo-labels $\hat{y}^{v}_t\in \mathbb{R}^C$:
\vspace{-1pt}
\begin{equation}
\hat{y}^{v}_t = \mathbbm{1}_{\{ \hat{z}^{v}_t > \theta^v\}} \odot y,\ 
\hat{z}^{v}_t = {E^{\textit{CLIP}}} \cdot {g^v_{L,t}}^\top, 
\end{equation}
where $y$ denotes the ground-truth video-level labels, $\mathbbm{1}_{\{\cdot\}}$ is the indicator function which returns a value of $1$ when the condition is true otherwise $0$, and $\odot$ denotes the element-wise product operation. The $\odot$ operation zeroes out the predictions of event classes absent in the video-level label.

A similar pseudo-label generation process is employed on the acoustic side. Raw waveforms of audio segments are first fed into CLAP's audio encoder and then into the pre-trained audio transformer. The event names of the classes in the target dataset for the AVVP task are filled in the caption template: ``This is the sound of $<$EVENT NAME$>$'' to generate textual event features: $E^{\textit{CLAP}} \in \mathbb{R}^{C\times d_1}$. Segment-level audio logits $\hat{z}^{a}_t \in \mathbb{R}^{C}$ and binary pseudo-labels $\hat{y}^{a}_t \in \mathbb{R}^C$ are then derived using class-wise thresholds $\theta^a\in \mathbb{R}^C$.

With binary segment-level pseudo-labels for both modalities $\hat{y}^{v}_t, \hat{y}^{a}_t$ and the predicted probabilities from the inference module (HAN) $\hat{p}^{v}_t, \hat{p}^{a}_t$ in place, the inference module can be trained using the binary cross-entropy loss as shown:
\vspace{-5pt}
\begin{equation}
\mathcal{L}_{hard} = \text{BCE}(p^v_t, \hat{y}^{v}_t) + \text{BCE}(p^a_t, \hat{y}^{a}_t).
\end{equation}

\subsection{Training with Pseudo-Label Uncertainty}
While pseudo-labels do provide additional supervision for better training of the inference module, they could potentially be noisy, leading to occasionally incorrect training signals.
To ameliorate this problem, we propose an uncertainty-weighted pseudo-label training scheme to improve the robustness of the learning process.
Instead of simply training with the binary pseudo-labels $\hat{y}^{v}_t, \hat{y}^{a}_t$, we leverage the confidence of the pseudo-label estimation module (associated with the predicted pseudo-label) to weigh the training signal for the inference module. This confidence score serves as a measure of the pseudo-label generation module's uncertainty of its prediction. This may be represented as:
\vspace{-5pt}
\begin{equation}
\hat{p}^{v}_t \!=\! \text{Sigmoid}(\hat{z}^{v}_t {-} \theta^v) \odot y; \,\ 
\hat{p}^{a}_t \!=\! \text{Sigmoid}(\hat{z}^{a}_t {-} \theta^a) \odot y.
\end{equation}
In other words, considering the visual pseudo-label generation pipeline as an example, the farther the logit $\hat{z}^{v}_t$ is from the threshold $\theta^v$, whether much lower or higher, the more confident the pseudo-label generation module is about the label it predicts (either approaching $0$ or $1$). 
Conversely, the closer the logit is to the threshold, the less the certainty about the correctness of the pseudo-labels (probabilities closer to $0.5$). An analogous explanation also holds for the audio pseudo-labels.
With the uncertainty-weighted pseudo-labels in place, the inference module (HAN) can be trained with the following uncertainty-weighted pseudo-label loss:
\vspace{-5pt}
\begin{equation}
\mathcal{L}_{soft} = \text{BCE}(p^v_t, \hat{p}^{v}_t) + \text{BCE}(p^a_t, \hat{p}^{a}_t).
\end{equation}

\subsection{Uncertainty-weighted Feature Mixup}
Due to the lack of full supervision for the weakly-supervised AVVP task, we explore the efficacy of additional regularization via self-supervision to help the models generalize better. Towards this end, prior pseudo-label generation approaches~\citep{wu2021exploring, rachavarapu2024weakly} often employ contrastive learning as a tool to better train the inference module. However, due to the inherent noise in the estimated pseudo-labels, positive samples and negative samples may be mislabeled, decreasing the effectiveness of the self-supervisory training. As an alternative, in this work, we explore the effectiveness of feature mixing, as a self-supervisory training signal for additional regularization. In this setting, we mixup the estimated features of any two segments, additively, and train the model to predict the union of the labels of the two segments.
However, since the labels in our setting are noisy, the mixed feature is assigned a label derived from a weighted sum of the uncertainty-weighted pseudo-labels of each of the two segment features. This is illustrated below:
\begin{align}
\bar{f}^{v}_{t_i, t_j}\!&= \lambda \Tilde{f}^{v}_{t_i} \!+\! (1\!-\!\lambda)\Tilde{f}^{v}_{t_j},~\ 
\bar{p}^{v}_{t_i, t_j}\!= \lambda \hat{p}^{v}_{t_i} \!+\! (1\!-\!\lambda)\hat{p}^{v}_{t_j} \\
\label{eq:mixup}
\bar{f}^{a}_{t_i, t_j}\!&= \lambda \Tilde{f}^{a}_{t_i} \!+\! (1{-}\lambda)\Tilde{f}^{a}_{t_j},~\ 
\bar{p}^{a}_{t_i, t_j}\!= \lambda \hat{p}^{a}_{t_i} \!+\! (1\!-\!\lambda)\hat{p}^{a}_{t_j},
\end{align}
where $\lambda \sim \text{Beta}(\alpha, \alpha)$ and $\alpha$ is a hyper-parameter controlling the Beta distribution, and $t_i$ and $t_j$ indicate two segment indices in a batch of video segments.

After mixing the uni-modal segment-level features, we pass them through the classifier of the inference module and apply the sigmoid function to the output, obtaining mixed segment-level event probabilities $p^{mix-v}_t$ and $p^{mix-a}_t$.
These are used to train the inference model with the uncertainty-aware mixup loss, as shown below:
\begin{equation}
\mathcal{L}_{mix} = \text{BCE}(p^{mix-v}_t, \bar{p}^{v}_t) + \text{BCE}(p^{mix-a}_t, \bar{p}^{a}_t).
\end{equation}

\subsection{Class-balanced Loss Re-weighting}
Besides the aforementioned challenges of the AVVP task, most of the events in the event set are absent in the pseudo-labels of any (segment of a) video (\ie, most event classes are negative events) and only a few events are present (\ie, positive events are much fewer in number). As a result, the model is dominated by the loss from the negative events. When trained without factoring in this bias, the classifier tends to overfit the negative labels and ignore the positive ones.
To address this class imbalance issue, we introduce a \textit{class-balanced loss re-weighting} strategy to re-balance the importance of the losses from the negative and positive events for the uncertainty-weighted pseudo-label loss. Specifically, the loss from the positive events is multiplied by a weight proportional to the frequency of the segments with the negative events in the pseudo-labels, while the loss from the negative events is multiplied by a weight proportional to the frequency of the segments with the positive events in the pseudo-labels, as shown below:
\vspace{-5pt}
\begin{multline}
\mathcal{L}_{w-soft} = \!\!\!\sum_{m\in \{v,a\}}\! w^m_{pos}\cdot y\cdot \text{BCE}(p^m_t, \hat{p}^{m}_t) ~+~  \\[-.2cm]
w^m_{neg}\cdot (1\!-\!y)\cdot \text{BCE}(p^m_t, \hat{p}^{m}_t),
\end{multline}

\vspace{-15pt}

\begin{align}
w^m_{pos}&=\frac{\sum_{i=1}^N \sum_{t=1}^T \sum_{c=1}^C (1-\hat{y}^{m}_{i,t,c})}{NTC}\times W, \\
w^m_{neg}&=\frac{\sum_{i=1}^N \sum_{t=1}^T \sum_{c=1}^C \hat{y}^{m}_{i,t,c}}{NTC},
\end{align}
where $N$ denotes the number of videos in the training set, and $W$ is a hyper-parameter.

In summary, the inference module is trained on the AVVP task with the proposed class-balanced re-weighting, applied to the uncertainty-weighted classification loss, and the uncertainty-weighted feature mixup loss, as shown below:
\vspace{-10pt}
\begin{align}
\mathcal{L}_{total} = \mathcal{L}_{w-soft} + \mathcal{L}_{mix} + \mathcal{L}_{video}.
\end{align}
\begin{table*}[t]
\centering
\caption{\textbf{Comparison with state-of-the-arts methods on the LLP dataset.} Best performances are in \textbf{bold}, second-best in \underline{underlined}.
}
\vspace{-6pt}
\label{table:sota_comparison}
\captionsetup{singlelinecheck = false}

\resizebox{0.8\textwidth}{!}{
\begin{tabular}{ r | c c c c c | c c c c c} 
\toprule
 \multirowcell{2}{\textbf{Method}} & \multicolumn{5}{c|}{\textbf{Segment-level}} & \multicolumn{5}{c}{\textbf{Event-level}} \\
{} & \textbf{A} & \textbf{V} & \textbf{AV} & \textbf{Type} & \textbf{Event} & \textbf{A} & \textbf{V} & \textbf{AV} & \textbf{Type} & \textbf{Event} \\

\midrule
{HAN~\citep{tian2020unified}} & 60.1 & 52.9 & 48.9 & 54.0 & 55.4 & 51.3 & 48.9 & 43.0 & 47.7 & 48.0 \\
{MA~\citep{wu2021exploring}} & 60.3 & 60.0 & 55.1 & 58.9 & 57.9 & 53.6 & 56.4 & 49.0 & 53.0 & 50.6 \\
{JoMoLD~\citep{cheng2022joint}} & 61.3 & 63.8 & 57.2 & 60.8 & 59.9 & 53.9 & 59.9 & 49.6 & 54.5 & 52.5 \\
{CMPAE~\citep{gao2023collecting}} & \underline{64.2} & 66.4 & 59.2 & 63.3 & 62.8 & 56.6 & 63.7 & 51.8 & 57.4 & 55.7 \\
{PoiBin~\citep{rachavarapu2023boosting}} & 63.1 & 63.5 & 57.7 & 61.4 & 60.6 & 54.1 & 60.3 & 51.5 & 55.2 & 52.3 \\
{VPLAN~\citep{zhou2023improving}} & 60.5 & 64.8 & 58.3 & 61.2 & 59.4 & 51.4 & 61.5 & 51.2 & 54.7 & 50.8 \\
{VALOR~\citep{lai2023modality}} & 61.8 & 65.9 & 58.4 & 62.0 & 61.5 & 55.4 & 62.6 & 52.2 & 56.7 & 54.2 \\
{LSLD~\citep{fan2023revisit}} & 62.7 & \underline{67.1} & 59.4 & 63.1 & 62.2 & 55.7 & \underline{64.3} & 52.6 & 57.6 & 55.2 \\
{PPL~\citep{rachavarapu2024weakly}} & \textbf{65.9} & 66.7 & \underline{61.9} & \underline{64.8} & \underline{63.7} & 57.3 & \underline{64.3} & \underline{54.3} & \underline{59.9} & \textbf{57.9} \\
{CoLeaf~\citep{sardari2025coleaf}} & \underline{64.2} & 64.4 & 59.3 & 62.6 & 62.5 & \underline{57.6} & 63.2 & 54.2 & 57.9 & 55.6 \\
{LEAP~\citep{zhou2024label}} & 62.7 & 65.6 & 59.3 & 62.5 & 61.8 & 56.4 & 63.1 & 54.1 & 57.8 & 55.0 \\
\midrule
{\name~(Ours)} & \underline{64.2} & \textbf{70.0} & \textbf{63.4} & \textbf{65.9} & \textbf{63.9} & \textbf{58.6} & \textbf{66.7} & \textbf{57.5} & \textbf{60.9} & \underline{57.4} \\

\bottomrule
\end{tabular}
}
\vspace{-10pt}
\end{table*}

\vspace{-20pt}
\section{Experiments}
We assess the performance of \name{} empirically across two challenging datasets and report its performance, comparing it with existing state-of-the-art approaches both quantitatively and qualitatively. Additionally, through multiple ablation studies, we bring out the effectiveness of the different elements of our proposed approach and the choices of different hyper-parameters. For additional details, ablation studies, and more qualitative results, we refer the reader to our supplementary material.

\subsection{Experimental Setup}
\paragraph{Datasets:}
We evaluate all competing methods on the \textit{Look, Listen, and Parse} (LLP) dataset~\citep{tian2020unified}, which is the principal benchmark dataset for the AVVP task. The dataset consists of $11,849$ video clips sourced from YouTube. Each clip is $10$ seconds long and represents one or more of $25$ diverse event categories, such as human activities, animals, musical instruments, and vehicles. The dataset is split into training, validation, and testing sets, following the official split~\citep{tian2020unified}: $10,000$ videos for training, $649$ videos for validation, and $1,200$ videos for testing. While the training set of this dataset is only associated with video-level labels of the events, the validation and testing split is labeled with segment-level event labels for evaluation purposes.
Additionally, to demonstrate the generalizability of our method, we conduct a similar study on the \textit{Audio Visual Event} (AVE) recognition dataset~\citep{tian2018audio}. 
The AVE dataset consists of $4,143$ video clips crawled from YouTube, each $10$ seconds long. It is split into $3,339$ videos for training, $402$ for validation, and $402$ for testing. It includes $29$ event categories encompassing human activities, animals, musical instruments, vehicles, and a ``background'' class (\ie, no event occurs). Unlike the LLP dataset, each video in the AVE dataset contains only one audio-visual event.
Here too, the training data is only provided with video-level labels while the validation and test splits are annotated with ground-truth event labels for each one-second segment, which either is a specific audio-visual event or ``background''.

\vspace{-5pt}
\paragraph{Metrics:}
For the LLP dataset, following the standard evaluation protocol~\citep{tian2020unified}, all models are evaluated using macro F1-scores calculated for the following event types: (i) audio-only (A), (ii) visual-only (V), and (iii) audio-visual (AV). Type@AV (Type) and Event@AV (Event) are two additional metrics that evaluate the overall performance of the model, where Type@AV is the mean of the F1-scores for the A, V, and AV events, while Event@AV is the F1-score of all events regardless of the modality in which they occur. Evaluations are conducted at both the segment-level and the event-level. At the segment-level, the model's predictions are compared with the ground truth on a per-segment basis. At the event-level, consecutive positive segments for the same event are grouped together as a single event. The F1-score is then computed using a mIoU threshold of $0.5$.
For the AVE dataset, we follow~\citet{tian2018audio} and use accuracy as the evaluation metric. An event prediction of a segment is considered correct if it matches the ground-truth label for that segment.

\vspace{-8pt}
\paragraph{Implementation Details:}
In line with prior work~\citep{tian2020unified}, each $10$-second video in both datasets is split into $10$ segments of one second each, where each segment contains $8$ frames. For the LLP dataset, pre-trained ResNet-152~\citep{he2016deep} and R(2+1)D-18~\citep{tran2018closer} are used to extract 2D and 3D visual features, respectively. The pre-trained VGGish~\citep{hershey2017cnn} network is used to extract features from the audio, sampled at 16KHz.
For the AVE dataset however, akin to prior work~\citep{lai2023modality}, we extract visual features from pre-trained CLIP and R(2+1)D while CLAP is used to embed the audio stream.  
For both datasets, we set the number of encoder blocks $L$ in the temporal-aware model to $5$, $\alpha$ for the Beta distribution in the feature mixup strategy to $1.7$, and $W$ in the class-balanced loss re-weighting step to $0.5$.
The pseudo-label generation modules and the inference model (HAN) are trained with the AdamW optimizer~\citep{loshchilovdecoupled}.
For improved performance on the AVE dataset, we replace $\hat{p}^{a}_{t_i}, \hat{p}^{a}_{t_j}$ in Eq.~\ref{eq:mixup} with $\lceil\hat{p}^{a}_{t_i}\rceil, \lceil\hat{y}^{a}_{t_j}\rceil$ and make corresponding modifications in the visual counterparts as well.

\vspace{-8pt}
\paragraph{Baselines:}
We demonstrate the effectiveness of \name{} by comparing against an extensive set of baselines.
For the LLP dataset, this includes video-level pseudo-label generation methods (MA~\citep{wu2021exploring}, JoMoLD~\citep{cheng2022joint}), segment-level pseudo-label generation methods (VALOR~\citep{lai2023modality}, LSLD~\citep{fan2023revisit}, PPL~\citep{rachavarapu2024weakly}), and the recently released works (CoLeaf~\citep{sardari2025coleaf}, LEAP~\citep{zhou2024label}).
On the other hand, for the AVE dataset, baseline approaches with publicly available implementation, which use the state-of-the-art feature backbones (akin to ours), such as HAN~\cite{tian2020unified} and VALOR~\cite{lai2023modality} are used.

\subsection{Results}
\paragraph{Comparison with Previous Methods on LLP:}
As shown in Table~\ref{table:sota_comparison}, \name{} surpasses previous methods, across almost all metrics. Notably, we achieve an F-score of \textbf{70.0} on the segment-level visual event, \textbf{65.9} on the segment-level Type@AV, and \textbf{66.7} on the event-level visual event metric.
This corresponds to a gain of $1.1\%$ on segment-level Type@AV F-score and a $1\%$ improvement on event-level Type@AV F-score, over our closest competitor PPL~\citep{rachavarapu2024weakly}. Of particular note, is the fact that our segment-level and event-level F-scores improve by more than \underline{$3\%$}, over PPL, for visual events. 
When compared to other recently published works, such as VALOR~\citep{lai2023modality}, CoLeaf~\citep{sardari2025coleaf} and LEAP~\citep{zhou2024label}, \name{} outperforms them by up to $3\%$ on both segment and event-level Type@AV F-score while gains on visual events are up to \underline{$5\%$} on segment-level F-score.

\begin{figure*}[t]
  \centering
  \includegraphics[width=0.9\textwidth]{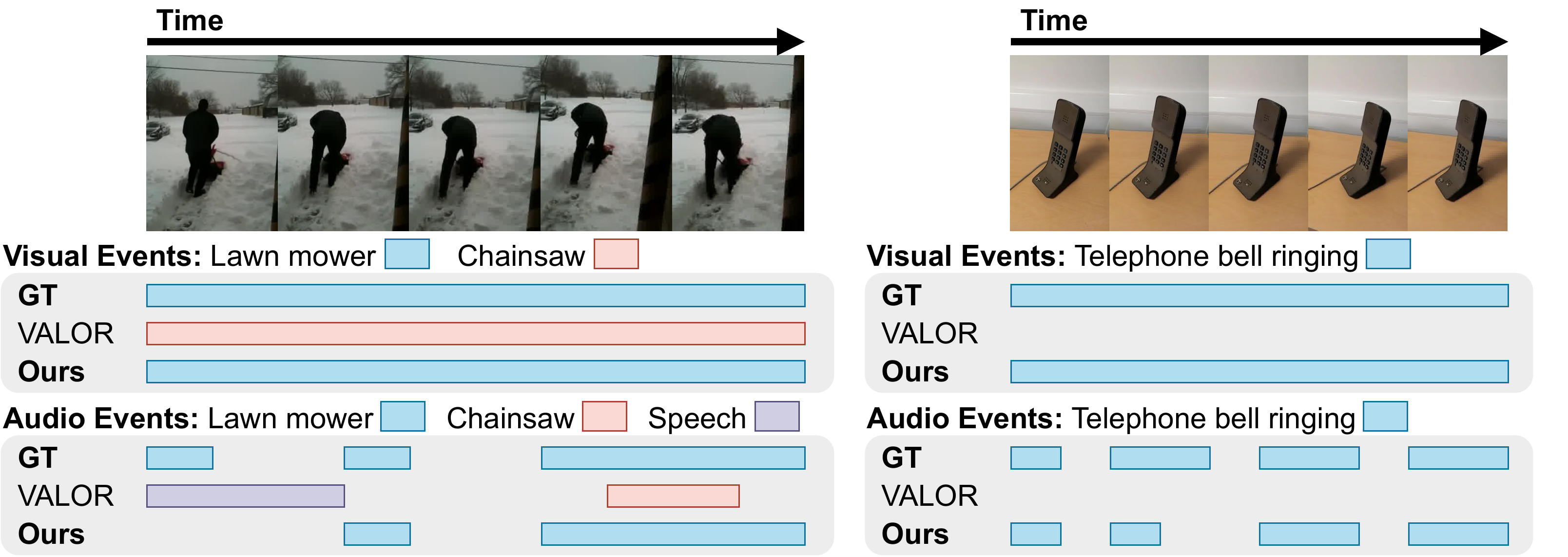}
  \caption{\textbf{Comparison between predictions by \name{} and competing AVVP methods on the LLP dataset.}
  ``GT'': ground truth.}
  \label{fig:qualitative_result}
  \vspace{-10pt}
\end{figure*}

These observations are consistent with our qualitative comparisons, as well. In the example on the left of Figure~\ref{fig:qualitative_result}, our model successfully recognizes and temporally localizes the lawn mower event visually, whereas VALOR~\citep{lai2023modality} (a recent state-of-the-art approach with publicly available implementation) misclassifies it as a chainsaw. Additionally, our model also accurately localizes the intermittent sound of the lawn mower. In contrast, VALOR not only misclassifies the sound of the lawn mower as that of a chainsaw but also incorrectly predicts that someone is talking in the video.
In the example on the right of Figure~\ref{fig:qualitative_result}, our model does not err in recognizing either the visual presence or the audio presence of the telephone, while VALOR fails to accurately predict the events in either modality.

\begin{table}[t]
\centering
\small
\caption{\textbf{Model performances on the AVE dataset.} CLIP, R(2+1)D-18, and CLAP are used as feature backbones.}
\renewcommand{\arraystretch}{0.7} 
\vspace{-6pt}
\resizebox{0.9\columnwidth}{!}{
    \begin{tabular}{c | c c c}
    \toprule
    Method   & HAN~\cite{tian2020unified}  & VALOR~\citep{lai2023modality} & \name{} (Ours) \\
    \midrule
    Acc.(\%) & 75.3 & 80.4 & \textbf{80.6} \\
    \bottomrule
    \end{tabular}
}
\label{tab:ave_exp}
\vspace{-5pt}
\end{table}

\vspace{-5pt}
\paragraph{Comparison with Previous Methods on AVE:}
To demonstrate the generalizability of our method, we evaluate \name{} on the AVE~\citep{tian2018audio} dataset and compare its performance with that of previous works. From Table~\ref{tab:ave_exp}, we observe that with the same backbone features, \name{} surpasses VALOR, our closest competitor, even on this small-scale dataset.

\vspace{-5pt}
\paragraph{Accuracy of the Generated Pseudo-Labels:}
\begin{table}[t]
\centering
\small
\caption{
\textbf{Accuracy of the generated pseudo-labels on LLP.}
}
\renewcommand{\arraystretch}{0.8} 
\vspace{-6pt}
\label{table:pseudo_label_reliability}
\captionsetup{singlelinecheck = false}

\resizebox{0.9\columnwidth}{!}{
    \begin{tabular}{ c | c c c c c } 
        \toprule
        \multirowcell{2}{Method} & \multicolumn{5}{c}{\textbf{Segment-level}} \\
        {} & \textbf{A} & \textbf{V} & \textbf{AV} & \textbf{Type} & \textbf{Event} \\
        
        \midrule
        {VALOR~\citep{lai2023modality}} & \textbf{80.5} & 61.7 & 55.7 & 66.0 & 74.6 \\ 
        {PPL~\citep{rachavarapu2024weakly}} & 61.7 & 61.8 & 57.5 & 60.6 & 59.4 \\
        \name{} (Ours) & 78.4 & \textbf{74.5} & \textbf{65.5} & \textbf{72.8} & \textbf{78.4} \\
        \bottomrule
    \end{tabular}
}
\vspace{-5pt}
\end{table}
To evaluate the efficacy of our pseudo-label generation pipeline, we compare the accuracy of our generated pseudo-labels against those obtained from competing methods (with publicly available implementation)~\citep{lai2023modality, rachavarapu2024weakly} on the test set of the LLP dataset. As shown in Table~\ref{table:pseudo_label_reliability}, our pre-trained temporally-dependent pseudo-label generation scheme generates more accurate segment-level pseudo-labels than previous methods, by up to \underline{$6\%$} on the segment-level Type@AV F-score, attesting to the advantages of factoring in inter-segment temporal dependencies.

\subsection{Ablation Study}
\begin{table}[t]
\centering
\caption{\textbf{Ablation study of the proposed components in \name.} ``Binary'' denotes training with binary pseudo-labels. ``Soft'' denotes training with uncertainty-weighted pseudo-labels.
}
\vspace{-5pt}
\label{table:ablation_study}
\captionsetup{singlelinecheck = false}

\resizebox{1.0\columnwidth}{!}{
    \begin{tabular}{ c@{~~}c | c@{~~}c | c@{~~~}c@{~~~}c@{~~~}c@{~~~}c } 
        \toprule
        \multirowcell{2}{Binary} & \multirowcell{2}{Soft} & \multirowcell{2}{Re-weight} & \multirowcell{2}{Mixup} & \multicolumn{5}{c}{\textbf{Segment-level}} \\
        {} & {} & {} & {} & \textbf{A} & \textbf{V} & \textbf{AV} & \textbf{Type} & \textbf{Event} \\
        
        \midrule
        {\cmark} & {} & {} & {} & 62.7 & 67.7 & 61.9 & 64.2 & 62.2 \\
        {} & {\cmark} & {} & {} & 63.0 & 68.3 & 61.8 & 64.4 & 62.8 \\
        \midrule
        {} & {\cmark} & {\cmark} & {} & 63.6 & 69.5 & 63.0 & 65.4 & 63.1 \\
        {} & {\cmark} & {} & {\cmark} & 63.9 & 69.0 & 62.8 & 65.2 & 63.4 \\
        {} & {\cmark} & {\cmark} & {\cmark} & \textbf{64.2} & \textbf{70.0} & \textbf{63.4} & \textbf{65.9} & \textbf{63.9} \\
        \bottomrule
    \end{tabular}
}
\vspace{-10pt}
\end{table}
To demonstrate the potency of the different elements of our proposed method, \name{}, we conduct ablation studies. In particular, the proposed uncertainty-weighted pseudo-label based training, the uncertainty-weighted feature mixup scheme, and the class-balanced loss re-weighting schemes are ablated. As shown in Table~\ref{table:ablation_study}, incorporating the uncertainty-weighted pseudo-label training step improves the segment-level Type@AV F-score by $2\%$, compared to using binary pseudo-labels. This demonstrates the benefit of accounting for the uncertainty in the pseudo-label estimation module. Moreover, sans the class-balanced loss re-weighting strategy, the model's performance is worse off by $1\%$ on the Type@AV F-score, revealing the erroneous bias in the model's prediction arising from a skew of the class distribution. On the other hand, introducing the uncertainty-weighted feature mixup results in a gain of $0.8\%$ on the Type@AV F-score, which underscores the importance of this self-supervised regularization.
\section{Conclusions}
In this work, we address the challenging task of weakly-supervised audio-visual video parsing (AVVP), which presents a two-fold challenge: (i) potential misalignment between the events of the audio and visual streams, and (ii) the lack of fine-grained labels for each modality.
We observe that by considering the temporal relationship between segments, our proposed method (\name{}) is able to provide more reliable pseudo-labels for better training of the inference module. In addition, by factoring in the uncertainty associated with these estimated pseudo-labels, regularizing the training process with a feature mixup strategy, and correcting for class imbalance, \name{} achieves state-of-the-art results on the LLP and AVE datasets.

\renewcommand{\thefigure}{A\arabic{figure}}
\renewcommand{\thetable}{A\arabic{table}}

{
    \small
    \bibliographystyle{ieeenat_fullname}
    \bibliography{main}
}

\clearpage
\setcounter{page}{1}
\maketitlesupplementary

We begin this supplementary document by expounding on the limitations of our proposed \name{} method. 
In the section that follows, we elaborate on the implementation details, compute environment used for implementation, and the training and inference times of our proposed method. Then, we quantitatively compare with VALOR on the LLP datset using better backbone features. In Section~\ref{sec:params}, we put forward studies showcasing the sensitivity of our method to the choice of the hyper-parameters $\alpha, W$, followed by ablation studies about the various design choices of our model. Finally, we end this document by providing some qualitative visualizations of the predictions obtained by our method versus competing baselines on both the LLP and the AVE datasets.

The following summarizes the supplementary materials we provide:
\begin{itemize}
    \item Limitations.
    \item Implementation details of \name{}.
    \item Details of our compute environment.
    \item Compute time analysis.
    \item Quantitative comparison using better backbone features.
    \item Studies on the sensitivity of \name{} to the choice of $\alpha, W$.
    \item The Scalability of \name{}.
    \item Ablation studies on the different design choices.
    \item Qualitative results of \name{} versus competing methods for the AVVP task.
\end{itemize}

\section{Limitations}
Although \name{} achieves state-of-the-art results on the AVVP task, compared to competing methods, it requires additional training data to pre-train the pseudo-label generation module (on which we train for about 80 epochs).

\section{Implementation Details of \name{}}
\label{sec:implement}

To assess the effectiveness of our method, in line with prior work~\citep{tian2020unified}, each $10$-second video in both the LLP~\cite{tian2020unified} and AVE~\cite{tian2018audio} datasets is split into $10$ segments of one second each, where each segment contains $8$ frames. The visual feature backbone for the LLP dataset is based on the ResNet-152~\citep{he2016deep} network (pre-trained on the ImageNet dataset~\cite{deng2009imagenet}) for extracting 2D-appearance features, and the R(2+1)D~\citep{tran2018closer} network (pre-trained on the Kinetics-400 dataset~\cite{kay2017kinetics}) for extracting features that capture the visual dynamics, respectively. The VGGish~\citep{hershey2017cnn} network, pre-trained on the AudioSet dataset~\cite{gemmeke2017audio}, is used to extract features from the audio, sampled at 16KHz.
For the AVE dataset however, akin to prior work~\citep{lai2023modality}, we extract visual features from pre-trained CLIP~\cite{radford2021learning} and R(2+1)D, while CLAP~\cite{wu2023large} is used to embed the audio stream.  
For both datasets, we set the number of encoder blocks $L$ of the transformers in the pseudo-label generation module to $5$, $\alpha$ for the Beta distribution in the feature mixup strategy to $1.7$, and $W$ in the class-balanced loss re-weighting step to $0.5$.
Both the pseudo-label generation modules and the inference modules are trained with the AdamW optimizer~\citep{loshchilovdecoupled}. 
To train the model, we employ a learning rate scheduling strategy that warms up the learning rate for the first $10$ epochs to its peak of $\expnumber{1}{-4}$ and then decays according to a cosine annealing schedule, to the minimum, which is set to $\expnumber{1}{-5}$ for the pseudo-label generation models and $\expnumber{5}{-6}$ for the inference model. We clip the gradient norm at $1.0$ during training. For the LLP dataset, the training batch size is set to $64$ and the total number of training epochs to $80$ for both models, while the same is set to $16$ and $80$ for the AVE dataset.

\begin{figure*}[h!]
    \centering
    \begin{minipage}[c]{0.48\textwidth}
        \centering
        \includegraphics[width=1.0\textwidth]{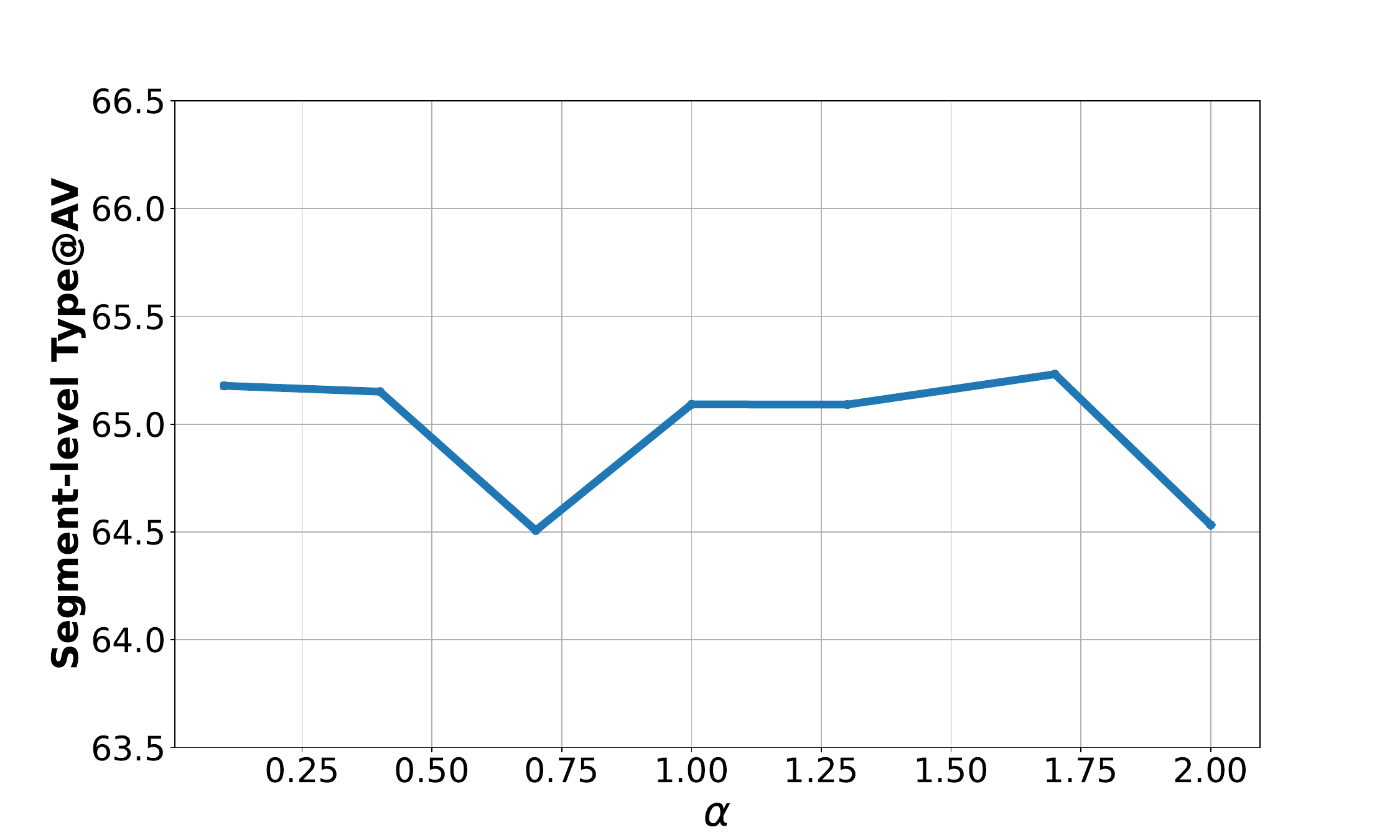}
        \vspace{-5pt}
        \caption{\textbf{Sensitivity of $\alpha$ in the uncertainty-weighted feature mixup on the LLP dataset.}}
        \label{fig:param_study_alpha}
    \end{minipage}
    \hfill
    \begin{minipage}[c]{0.48\textwidth}
        \centering
        \includegraphics[width=1.0\textwidth]{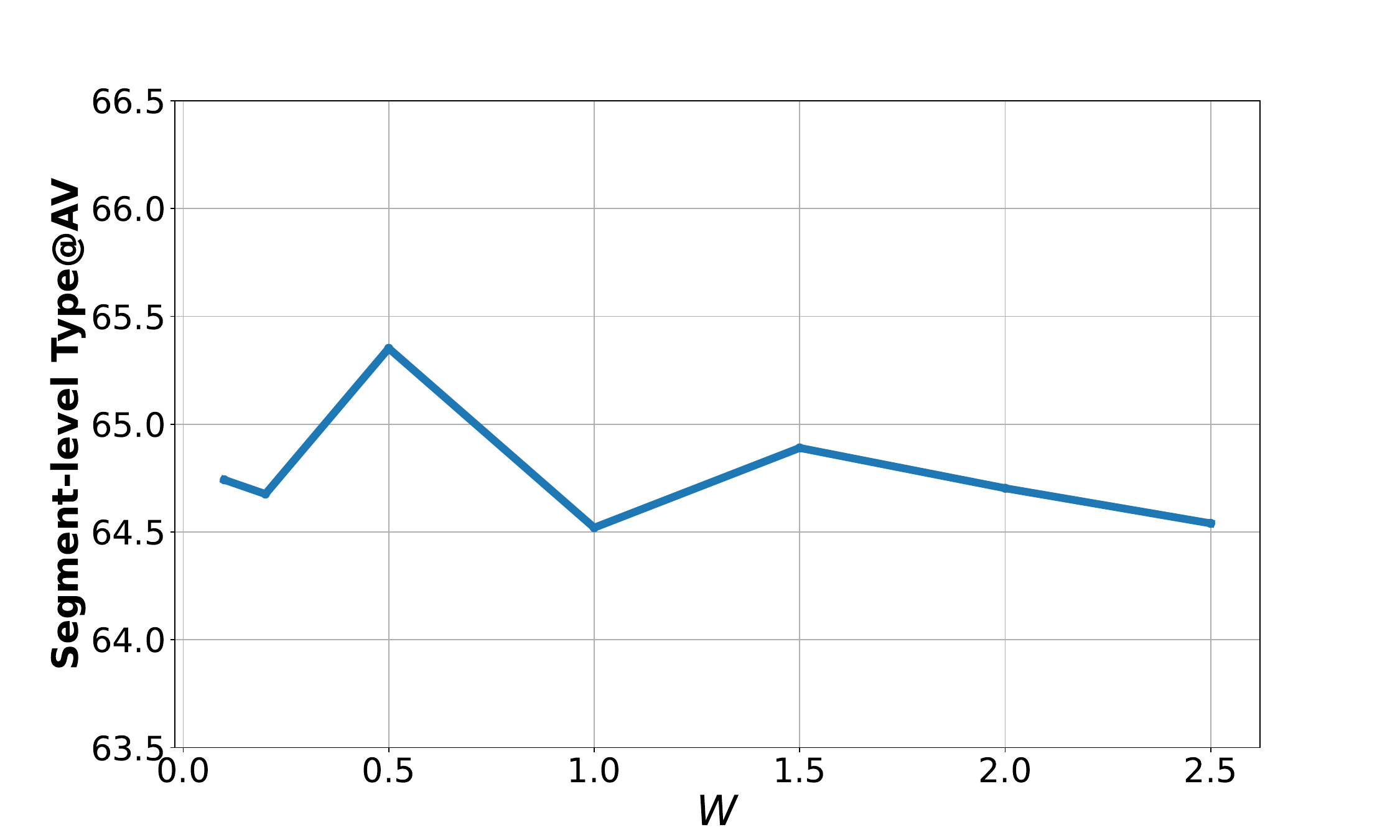}
        \vspace{-5pt}
        \caption{\textbf{Sensitivity of $W$ in the class-balanced re-weighting on the LLP dataset.}}
        \label{fig:param_study_weight}
    \end{minipage}
\end{figure*}

\section{Details of Compute Environment}
\label{sec:compute_infra}
Our model is trained on a desktop computer with an Intel Core i7 CPU, with 32GB RAM, and a \underline{single} NVIDIA RTX 3090 GPU.

\section{Analysis of Compute Time}
\label{sec:compute_anal}
\begin{table}[t]
\centering
\caption{\textbf{Compute time analysis on the LLP dataset.} ``Inference Time'' denotes the time to evaluate all testing data.}
\vspace{-5pt}
\resizebox{1.0\columnwidth}{!}{
    \begin{tabular}{c | c c}
    \toprule
    \textbf{Method}  & Training Time per Epoch & Inference Time  \\
    \midrule
    CoLeaf~\cite{sardari2025coleaf} & 25 sec & 24 sec \\
    \midrule
    \name{} (Ours) & 24 sec & 20 sec \\
    \bottomrule
    \end{tabular}
}
\label{tab:compute_time_analysis}
\end{table}
For a more holistic understanding of the performance of our method, we compare its training and inference times with the most recently published approach for the AVVP task, viz. CoLeaf~\cite{sardari2025coleaf} on the LLP dataset~\cite{tian2020unified}.
The results of this study are shown in Table~\ref{tab:compute_time_analysis}. We see that our method's runtime performances are comparable with those of competing approaches, with notable inference time gains over the CoLeaf method~\cite{sardari2025coleaf}.

\begin{table*}[ht!]
\centering
\caption{\textbf{Comparison with VALOR on the LLP dataset.} $^\dagger$~denotes using CLIP and CLAP features as input to the HAN model.}
\vspace{-5pt}
\label{table:better_feat_comparison}
\captionsetup{singlelinecheck = false}

\resizebox{0.9\textwidth}{!}{
\begin{tabular}{ r | c c c c c | c c c c c} 
\toprule
 \multirowcell{2}{\textbf{Method}} & \multicolumn{5}{c|}{\textbf{Segment-level}} & \multicolumn{5}{c}{\textbf{Event-level}} \\
{} & \textbf{A} & \textbf{V} & \textbf{AV} & \textbf{Type} & \textbf{Event} & \textbf{A} & \textbf{V} & \textbf{AV} & \textbf{Type} & \textbf{Event} \\

\midrule
{VALOR$^\dagger$ [17]} & 68.1 & 68.4 & 61.9 & 66.2 & 66.8 & 61.2 & 64.7 & 55.5 & 60.4 & 59.0 \\
{\name$^\dagger$~(Ours)} & \textbf{68.9} & \textbf{72.3} & \textbf{65.6} & \textbf{68.9} & \textbf{68.3} & \textbf{63.5} & \textbf{68.7} & \textbf{59.6} & \textbf{63.9} & \textbf{62.4} \\

\bottomrule
\end{tabular}
}
\end{table*}
\section{Quantitative Comparison Using Better Backbone Features}
We also quantitatively compare our proposed approach with VALOR on the LLP dataset using better backbone features, \ie CLIP and CLAP as visual and audio feature backbones. As shown in Table~\ref{table:better_feat_comparison}, \name{} outperforms VALOR with $2.7$ F-score improvement in segment-level Type@AV and $3.5$ F-score improvement in event-level Type@AV.

\section{Sensitivity to the Choice of $\alpha$ and $W$}
\label{sec:params}
To gain a better understanding of the effect of the choice of hyper-parameters on our model's performance, we evaluate the sensitivity of our model to the choice $\alpha$ in the uncertainty-weighted feature mixup and $W$ in the class-balanced loss re-weighting strategy.
When $\alpha$ is adjusted, class-balanced loss re-weighting is not applied. As shown in Figure~\ref{fig:param_study_alpha}, for the LLP dataset, when $\alpha$ increases from $0.1$ to $2.0$, segment-level Type@AV F-score first decreases to $64.5$, then rises to a peak of $65.2$ at $\alpha=1.7$, and subsequently declines back to $64.5$.
On the other hand, Figure~\ref{fig:param_study_weight} illustrates the effect of varying $W$ on the segment-level Type@AV F-score. The F-score reaches its maximum value of $65.3$ when $W=0.5$ and decreases as $W$ becomes larger. When $W$ is adjusted, the uncertainty-weighted feature mixup is not applied. These observations point towards the robustness of our model to the precise choice of these hyper-parameters. We observe similar trends for the AVE dataset as well. Hence, for best results, we select $\alpha=1.7$ and $W=0.5$ in all our experiments for both datasets.

\section{The Scalability of \name}
To evaluate the scalability of \name, we train the inference model (HAN) with less training data (Table~\ref{tab:different_class_number}) as well as fewer event classes (Table~\ref{tab:different_data_amount}) on the LLP dataset by removing the training videos or event classes randomly. Even with only $60\%$ of the training data, \name{} exhibits competitive performance. Moreover, \name{} shows a consistent performance lead against VALOR [17], irrespective of the number of event classes, with no change in training strategy or the core model structure.
\begin{table}[t!]
  \centering
  \label{tab:scalability}
  \caption{\textbf{The scalability of \name.}}

  \vspace{-5pt}
  \begin{subtable}[t]{1.0\columnwidth}
    \centering
    \caption{Training with different amounts of data.}
    \label{tab:different_data_amount}
    \begin{tabular}{ c | c@{~~~}c@{~~~}c@{~~~}c@{~~~}c} 
        \toprule
         \multirowcell{2}[0pt][c]{Training Data\\Ratio} & \multicolumn{5}{c}{\textbf{Segment-level}} \\
        {} & \textbf{A} & \textbf{V} & \textbf{AV} & \textbf{Type} & \textbf{Event} \\
        \midrule
        {100\%} & \textbf{64.2} & \textbf{70.0} & \textbf{63.4} & \textbf{65.9} & \textbf{63.9} \\
        {80\%}  & 63.4 & 69.2 & 62.5 & 65.0 & 63.0 \\
        {60\%}  & 63.4 & 68.6 & 62.4 & 64.8 & 62.8 \\
        \bottomrule
    \end{tabular}
  \end{subtable}

  \vspace{10pt}

  \begin{subtable}[t]{1.0\columnwidth}
    \centering
    \caption{Training with different number of classes.}
    \label{tab:different_class_number}
    \begin{tabular}{ c | c@{~~~}c@{~~~}c}
        \toprule
        & \multicolumn{3}{c}{\textbf{Segment-level Type F-score}} \\[-2pt]
        \midrule
        \multirowcell{2}[0pt][c]{Number of Classes} & 
        \multirowcell{2}[0pt][c]{25\\[-2pt](all events)} &
        \multirowcell{2}[0pt][c]{20} & 
        \multirowcell{2}[0pt][c]{15} \\[-2pt]
        {} & & & \\
        \midrule
        {VALOR [17]} & 62.0 & 65.9 & 66.6 \\
        {\name (Ours)} & 65.9 & 71.4 & 68.4 \\[-1pt]
        \bottomrule
    \end{tabular}
  \end{subtable}
\vspace{-10pt}
\end{table}

\begin{table*}[ht!]
\centering
\caption{\textbf{Ablation study reported on all metrics.} ``Binary'' denotes training with binary pseudo-labels. ``Soft'' denotes training with uncertainty-weighted pseudo-labels.}
\vspace{-5pt}
\label{table:full_ablation}
\captionsetup{singlelinecheck = false}

\resizebox{\textwidth}{!}{
    \begin{tabular}{ c@{~~}c | c@{~~}c | c@{~~~}c@{~~~}c@{~~~}c@{~~~}c | c@{~~~}c@{~~~}c@{~~~}c@{~~~}c} 
        \toprule
        \multirowcell{2}{Binary} & \multirowcell{2}{Soft} & \multirowcell{2}{Re-weight} & \multirowcell{2}{Mixup} & \multicolumn{5}{c|}{\textbf{Segment-level}} & \multicolumn{5}{c}{\textbf{Event-level}} \\
        {} & {} & {} & {} & \textbf{A} & \textbf{V} & \textbf{AV} & \textbf{Type} & \textbf{Event} & \textbf{A} & \textbf{V} & \textbf{AV} & \textbf{Type} & \textbf{Event} \\
        
        \midrule
        {\cmark} & {} & {} & {} & 62.7 & 67.7 & 61.9 & 64.2 & 62.2 & 56.9 & 64.9 & 56.6 & 59.5 & 55.8 \\
        {} & {\cmark} & {} & {} & 63.0 & 68.3 & 61.8 & 64.4 & 62.8 & 56.9 & 65.2 & 55.9 & 59.3 & 56.1 \\
        \midrule
        {} & {\cmark} & {\cmark} & {} & 63.6 & 69.5 & 63.0 & 65.4 & 63.1 & 57.9 & 66.4 & 57.0 & 60.4 & 56.9 \\
        {} & {\cmark} & {} & {\cmark} & 63.9 & 69.0 & 62.8 & 65.2 & 63.4 & 57.7 & 65.6 & 56.3 & 59.9 & 56.8 \\
        {} & {\cmark} & {\cmark} & {\cmark} & \textbf{64.2} & \textbf{70.0} & \textbf{63.4} & \textbf{65.9} & \textbf{63.9} & \textbf{58.6} & \textbf{66.7} & \textbf{57.5} & \textbf{60.9} & \textbf{57.4} \\
        \bottomrule
    \end{tabular}
}
\end{table*}
\section{Ablation Studies}
\label{sec:ablation}

\begin{table}[ht!]
\centering
\small
\caption{\textbf{Ablation study of uncertainty-weighted mixup in Eq. 14 and Eq. 15. on the AVE dataset.}}
\vspace{-5pt}
\resizebox{0.8\columnwidth}{!}{
    \begin{tabular}{c | c c}
    \toprule
    Method  & $\hat{p}^{a}_{t}, \hat{p}^{v}_{t}$ & $\lceil\hat{p}^{a}_{t}\rceil, \lceil\hat{p}^{v}_{t}\rceil$  \\
    \midrule
    Acc.(\%) & 80.3 & \textbf{80.6} \\
    \bottomrule
    \end{tabular}
}
\label{tab:ave_ablation}
\end{table}
\paragraph{Ablation Study on All Metrics:}
In Table~\ref{table:full_ablation}, we report the ablation study on all metrics for a more complete understanding. Coupled with our proposed class-balanced re-weighting strategy, the HAN model improves from $59.3$ to $60.4$ for the event-level Type@AV. On the other hand, by introducing the proposed uncertainty-aware mixup strategy, the event-level Type@AV increases from $59.3$ to $60$.

\paragraph{Ablation Study of the Uncertainty-weighted Mixup on the AVE Dataset:}
As shown in Table~\ref{tab:ave_ablation}, our experiments reveal that using $\lceil\hat{p}^{a}_{t}\rceil$ and $\lceil\hat{p}^{v}_{t}\rceil$ as the segment-level pseudo labels instead of $\hat{p}^{a}_{t}$ and $\hat{p}^{v}_{t}$ for the uncertainty-weighted feature mixup strategy, in Eq. 14 and Eq. 15. in the main paper, results in a slightly better performance on the AVE dataset.

\section{Qualitative Results}
\begin{figure*}[t!]
  \centering
  \includegraphics[width=0.85\textwidth]{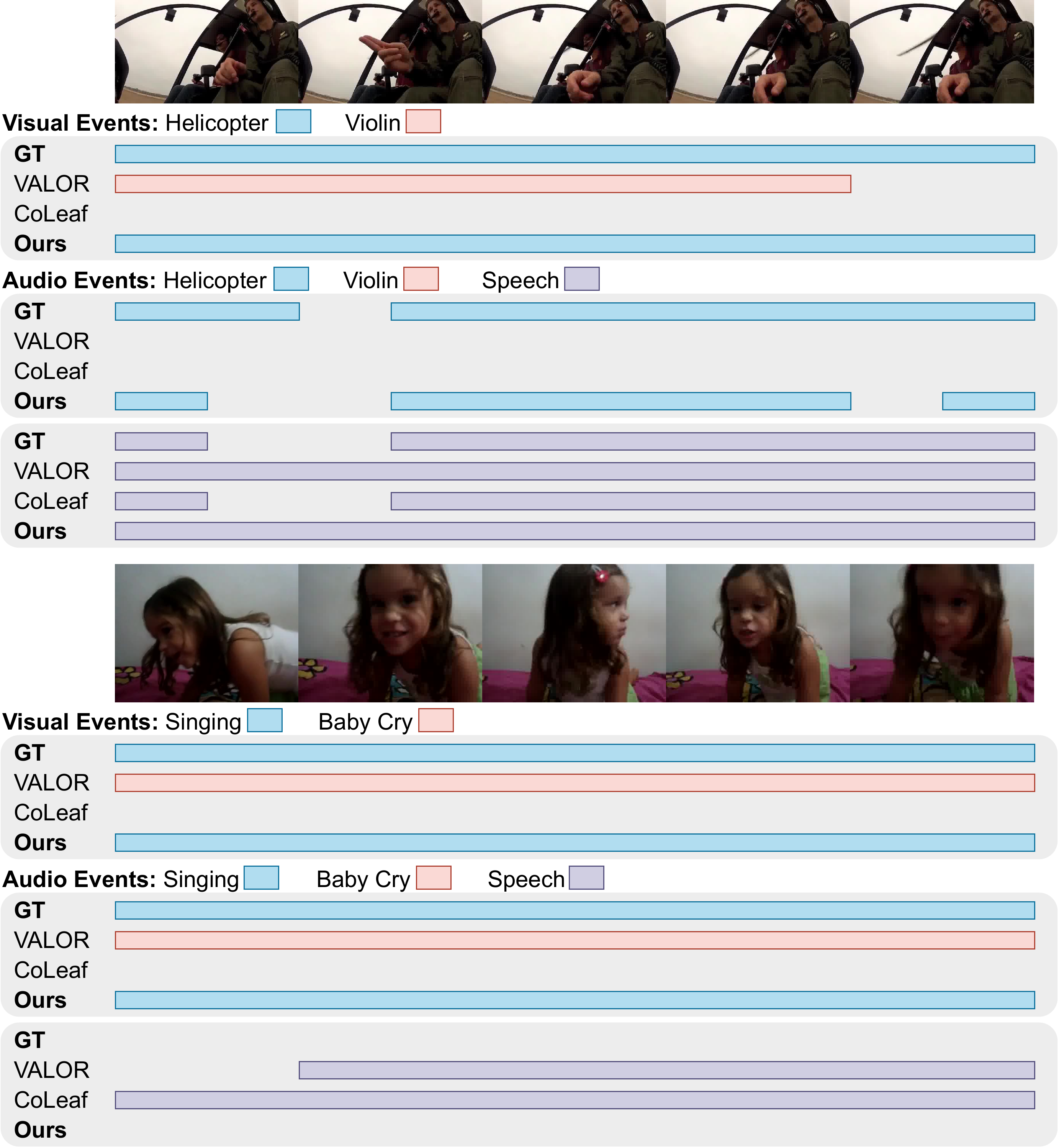}
  \caption{\textbf{Comparison between predictions by \name{} and competing AVVP methods on the LLP dataset.}
  ``GT'': ground truth.}
  \label{fig:supp_llp_qualitative_1}
\end{figure*}

\begin{figure*}[t]
  \centering
  \includegraphics[width=0.75\textwidth]{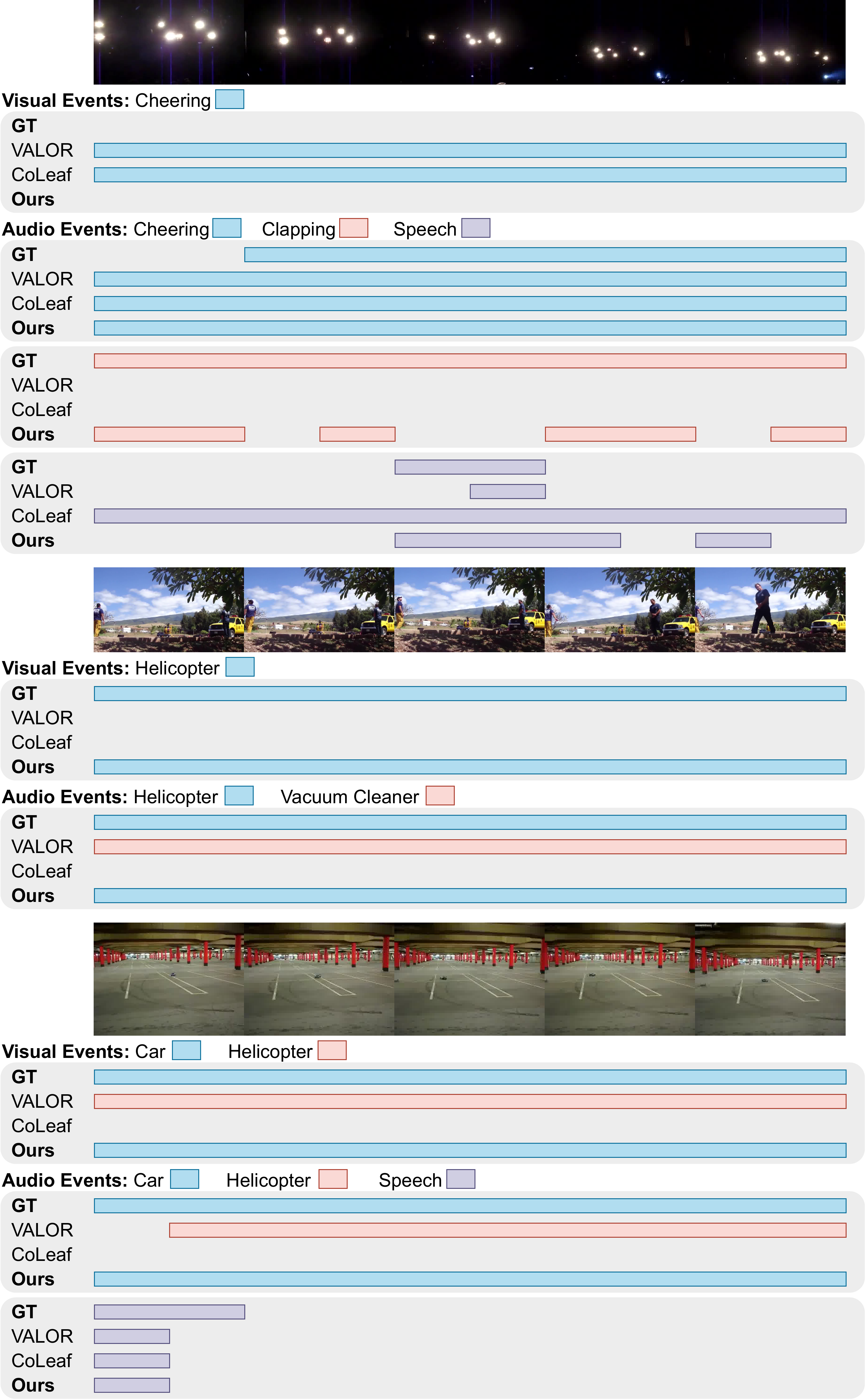}
  \caption{\textbf{Comparison between predictions by \name{} and competing AVVP methods on the LLP dataset.}
  ``GT'': ground truth.}
  \label{fig:supp_llp_qualitative_2}
\end{figure*}
\begin{figure*}[t]
  \centering
  \includegraphics[width=0.9\textwidth]{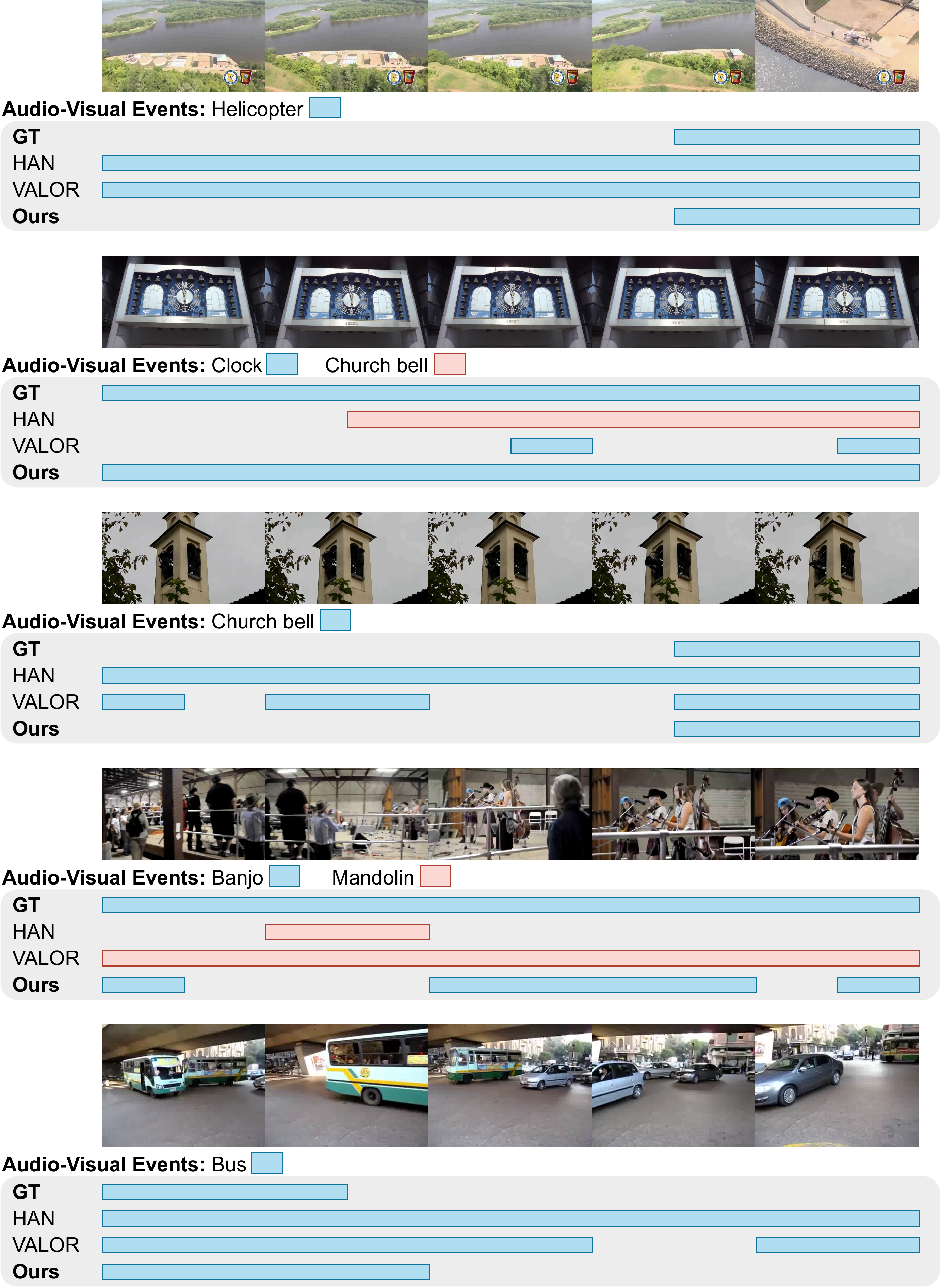}
  \caption{\textbf{Qualitative comparison between predictions by \name{} and previous methods on the AVE dataset.}
  ``GT'': ground truth.}
  \label{fig:supp_ave_qualitative_results}
\end{figure*}

Figures~\ref{fig:supp_llp_qualitative_1}, ~\ref{fig:supp_llp_qualitative_2} show event predictions of our method versus competing baselines on sample videos from the LLP dataset~\cite{tian2020unified}. Figure~\ref{fig:supp_ave_qualitative_results} shows the same, for sample videos on the AVE dataset~\cite{tian2018audio}. As is evident from the figures, we see consistently accurate event-label predictions across different videos, while also generally accurately localizing them, the same is not the case for the baseline approaches. This feature is particularly prominent for instance, for the visual event classes in the first video in Figure~\ref{fig:supp_llp_qualitative_2}, or the audio-visual events in the second video example in Figure~\ref{fig:supp_ave_qualitative_results}. However, there remain challenging scenarios where almost all methods struggle, such as the audio events in the first video example in Figure~\ref{fig:supp_llp_qualitative_2}, which we hope to address going forward.

\end{document}